\documentclass[10pt,twocolumn,letterpaper]{article}

\usepackage[pagenumbers]{cvpr} 

\usepackage[dvipsnames]{xcolor}

\usepackage{graphicx}
\usepackage{amsmath}
\usepackage{amssymb}
\usepackage{booktabs}
\usepackage{multirow}
\usepackage{adjustbox}
\usepackage{wrapfig}
\usepackage{multicol}
\usepackage{lipsum}
\usepackage{stfloats}
\usepackage[accsupp]{axessibility}

\usepackage{amsmath}
\usepackage{amssymb}
\usepackage{paralist}
\usepackage{textcomp}
\usepackage{placeins}

\usepackage{xcolor}
\definecolor{mylightgray}{gray}{0.5}
\definecolor{mydarkgreen}{RGB}{0, 150, 0}
\definecolor{mydarkred}{RGB}{200, 0, 0}
\definecolor{myblue}{RGB}{0, 0, 255}
\definecolor{myorange}{RGB}{255, 100, 0}

\usepackage[dvipsnames]{xcolor}

\definecolor{cvprblue}{rgb}{0.21,0.49,0.74}
\usepackage[pagebackref,breaklinks,colorlinks,citecolor=cvprblue]{hyperref}

\def\paperID{*****} 
\def\confName{CVPR}
\def\confYear{2024}

\title{Q-GroundCAM: Quantifying Grounding in Vision Language Models \\ via GradCAM}

\author{Navid Rajabi and Jana Ko{\v{s}}eck{\'a}\\
George Mason University\\
{\tt\small \{nrajabi, kosecka\}@gmu.edu}
}

\begin{document}
\maketitle

\begin{abstract}
Vision and Language Models (VLMs) continue to demonstrate remarkable zero-shot (ZS) performance across various tasks. However, many probing studies have revealed that even the best-performing VLMs struggle to capture aspects of compositional scene understanding, lacking the ability to properly ground and localize linguistic phrases in images. Recent VLM advancements include scaling up both model and dataset sizes, additional training objectives and levels of supervision, and variations in the model architectures. To characterize the grounding ability of VLMs, such as phrase grounding, referring expressions comprehension, and relationship understanding, Pointing Game has been used as an evaluation metric for datasets with bounding box annotations. In this paper, we introduce a novel suite of quantitative metrics that utilize GradCAM activations to rigorously evaluate the grounding capabilities of pre-trained VLMs like CLIP, BLIP, and ALBEF. These metrics offer an explainable and quantifiable approach for a more detailed comparison of the zero-shot capabilities of VLMs and enable measuring models' grounding uncertainty. This characterization reveals interesting tradeoffs between the size of the model, the dataset size, and their performance.
\end{abstract}
\begin{figure}[!t]
  \centering
  \includegraphics[scale=0.29]{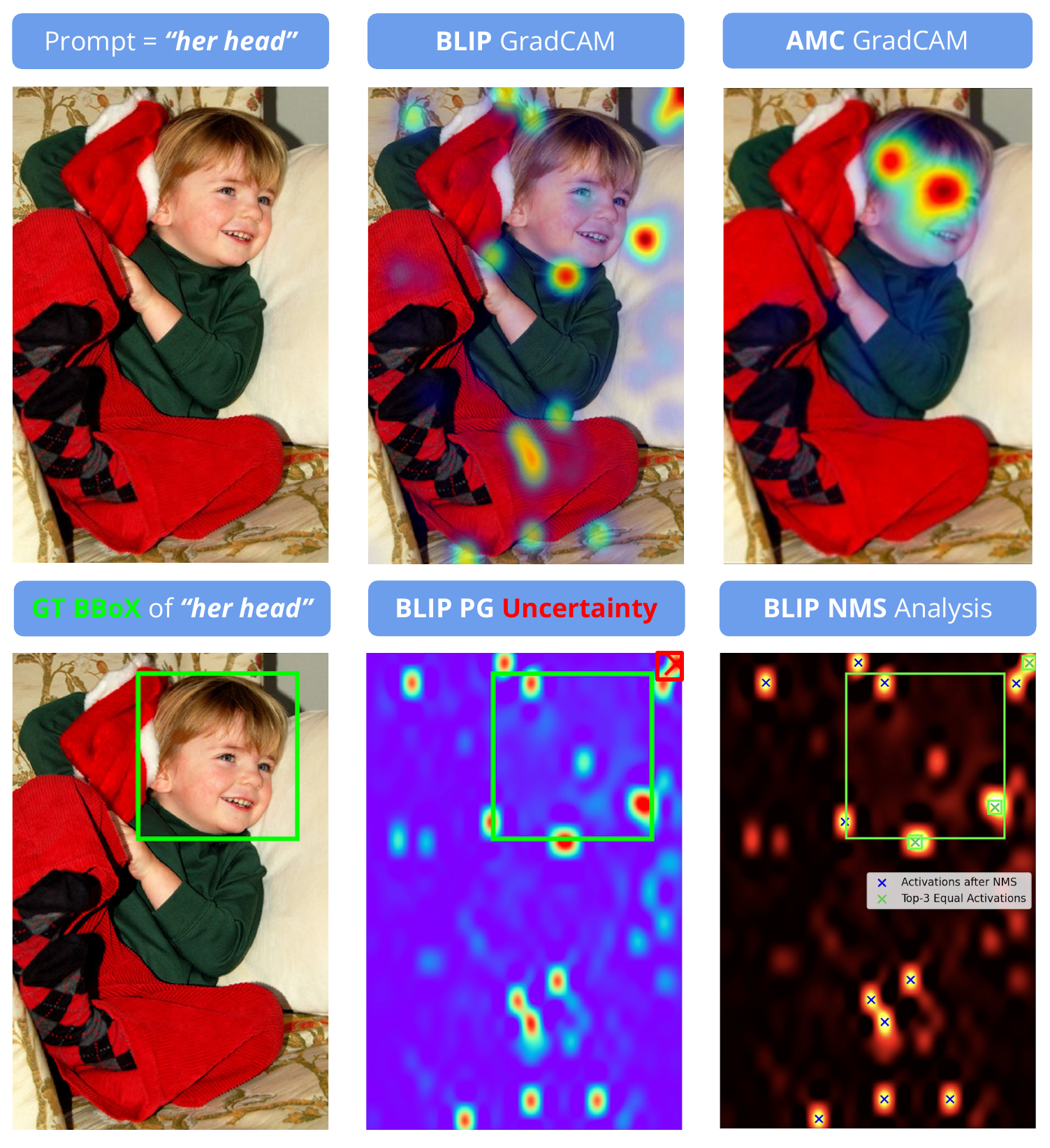}
   \caption{Uncertainty in Pointing Game (PG) accuracy, when having multiple top-\textit{k} identical activations with \textit{inconsistent} PG binary labels (\textbf{Scenario 1}). As depicted in the bottom-right figure, three top high-confidence activations exist, each with a value of \texttt{1.0}, after our NMS analysis. One falls \textit{outside} the bounding box, one \textit{inside}, and one at the \textit{border}. In these cases, PG lacks any additional clues or heuristics to determine which one to select.}
   \label{fig:introSenario1}
\end{figure}
\begin{figure*}[!t]
  \centering
  \includegraphics[width=1.\linewidth]{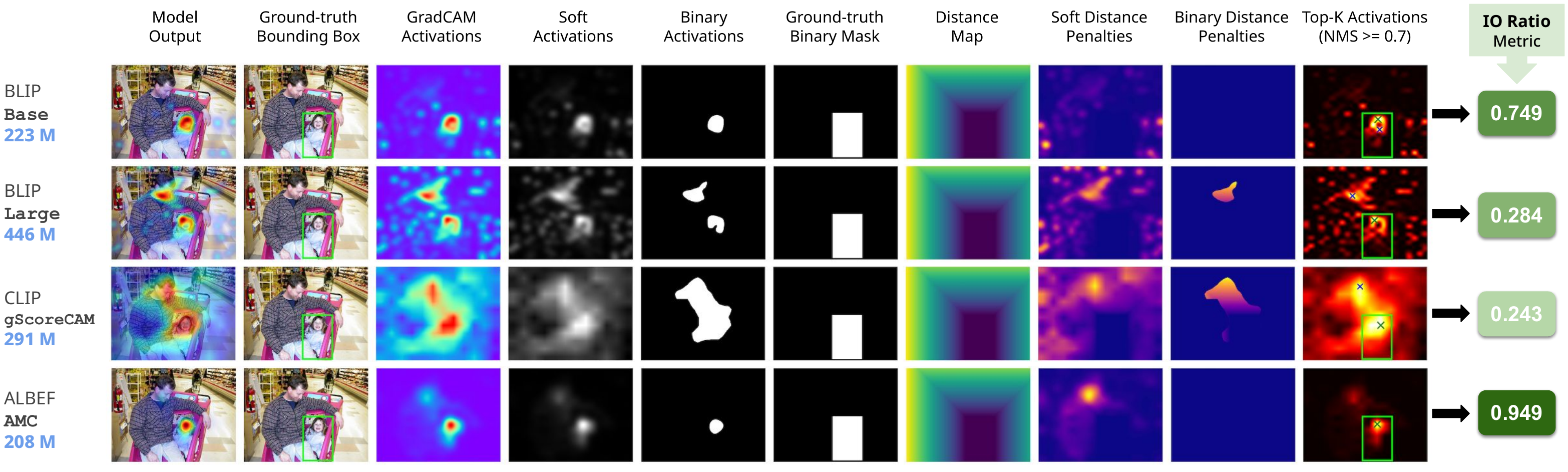}
   \caption{Given the \textit{\textbf{"his daughter"}} prompt, PG returns the same accuracy of \texttt{1} for all four model outputs in a \textit{discrete} manner, and overlooks the differences in their holistic grounding qualities (\textbf{Scenario 2}). On the other hand, our $\mathbf{IO}_{ratio}$ metric can differentiate and rank them in a more \textit{explainable} and \textit{continuous} manner by quantifying them each as a single normalized value between \texttt{0} and \texttt{1}.}
   \label{fig:2ndFigScenario2}
\end{figure*}
\section{Introduction}
\label{sec:intro}
Foundational Vision and Language Models (VLMs) have demonstrated impressive performance in various vision and language tasks, including visual question answering (VQA), retrieval, image-text matching, referring expression comprehension, or captioning. For instance, CLIP has become a widely-used backbone in various applications, ranging from open-vocabulary object localization \& referring expression grounding~\cite{chen2022gscorecam, reclip}
, to in-context learning in generative Multimodal Large Language Models (MLLMs) such as LLaVa~\cite{liu2023llava}, and Embodied AI for navigation-related tasks~\cite{huang2023vlmaps}.
Despite that, state-of-the-art VLMs still struggle to capture aspects of compositional scene understanding and lack proper grounding for noun phrases, verbs, or relations~\cite{svo, bag-of-words, winoground}. Earlier model architectures localize objects of interest by predicting bounding box locations and use traditional IoU evaluation metrics with respect to ground truth datasets~\cite{gpv, kamath2021mdetr, dou2022coarsefiber}. Recent MLLMs forgo the bounding box prediction and tackle the location prediction on the language side~\cite{liu2023llava, peng2023kosmos2}.  
The ability to ground and localize the concepts continues to be of central interest~\cite{amc}, and GradCAM visualizations~\cite{GradCam} or Pointing Game accuracy~\cite{pointinggameinit} are often used to quantify models' grounding ability.
Pointing Game (PG) considers grounding successful if the highest GradCAM activation falls inside the ground truth bounding box \cite{gupta2020contrastive}. While Pointing Game has been utilized effectively in characterization of grounding performance~\cite{datta2019align2ground, akbari2019multi, gupta2020contrastive, arbelle2021detector, wang2021improving, amc}, it offers only coarse 0/1 characterization, which is susceptible to spurious local maxima or existence of multiple maxima, and does not capture well the confidence in the grounded concept.
The examples in Figures~\ref{fig:introSenario1} \&~\ref{fig:2ndFigScenario2} would be incorrectly handled by the Pointing Game approach due to the following scenarios:
\begin{compactitem}
    \item \textbf{Scenario 1:} Highly confident activations, some in and some outside of the bounding box, are predicted by the model as in Figure~\ref{fig:introSenario1}. In these cases, the Pointing Game must randomly pick the local maximum. 
    We refer to this as $\mathrm{PG}_{Uncertainty}$ metric and report the total number of cases where this type of uncertainty occurs for each dataset in Table~\ref{tab:benchmarking}.
    \item \textbf{Scenario 2:} Pointing Game does not consider the spurious activation predictions outside the ground-truth bounding box when comparing the performance of different models on the same instance. As shown in Figure~\ref{fig:2ndFigScenario2}, while PG returns the label of \texttt{1} for all four models, our $\mathrm{IO}_{ratio}$ metric provides a \underline{finer-grained} characterization of grounding quality. 
\end{compactitem}
In order to provide a finer-grained characterization of the model's grounding ability, we propose a set of metrics to (1) compute the similarity between the $\mathrm{GradCAM}$ activation maps and ground-truth binary mask, (2) reward the activations inside while penalizing the \textit{spurious} activations outside of the ground-truth bounding box, and (3) measure the \textbf{Scenario 1} uncertainty. Our inspiration stems from the Attention Mask Consistency loss (AMC) introduced in \cite{amc} which is designed to \textcolor{mydarkgreen}{\textbf{maximize}} the activations \textcolor{mydarkgreen}{\textbf{inside}} and \textcolor{mydarkred}{\textbf{minimize}} them \textcolor{mydarkred}{\textbf{outside}} the ground-truth bounding box, used in the pre-training of ALBEF~\cite{albef}. 
Since our metrics are applicable to a wide range of VLMs, including both patch-based ViT~\cite{dosovitskiy2020imagevit} and CNN-based vision transformer encoders~\cite{clip, chen2022gscorecam}, they can thoroughly assess the grounding ability of VLMs, even without the bounding box prediction mechanism.
The proposed metrics can distinguish nuances in the comparison of models' grounding abilities that PG cannot capture.
We will use these metrics to evaluate four state-of-the-art VLMs ($\mathrm{BLIP}_{base}$ \cite{blip}, $\mathrm{BLIP}_{large}$ \cite{blip}, $\mathrm{CLIP}$ $\mathrm{gScoreCAM}$~\cite{clip, chen2022gscorecam}, and the $\mathrm{AMC}$ \cite{amc} variation of $\mathrm{ALBEF}$ \cite{albef}) on a wide spectrum of grounding tasks\footnote{We release the code at \href{https://github.com/NavidRajabi/Q-GroundCAM}{Github.com/NavidRajabi/Q-GroundCAM}.}.

\begin{table*}[!t] 
\scriptsize
\centering
\setlength{\tabcolsep}{6pt}
\renewcommand{\arraystretch}{1.0}
{
\scriptsize
\begin{tabular}{@{}llcccccc|c|cc@{}}
\toprule
\multirow{2}{*}{\small\textbf{Dataset}} & \multirow{2}{*}{\small\textbf{Model}} & \multicolumn{2}{c}{$\mathbf{IoU}$ $\mathbf{\uparrow}$} & \multicolumn{2}{c}{$\mathbf{Dice}$ $\mathbf{\uparrow}$} & \multicolumn{2}{c}{$\mathbf{WDP}$ $\mathbf{\downarrow}$} & \multicolumn{1}{c}{\textbf{$\mathbf{IO}_{ratio}$ $\uparrow$}} & \multicolumn{2}{c}{$\mathbf{PG}$} \\

\cmidrule(lr){3-4} \cmidrule(lr){5-6} \cmidrule(lr){7-8} \cmidrule(lr){9-9} \cmidrule(lr){10-11}

 &  & \textbf{Soft} & \textbf{Binary} & \textbf{Soft} & \textbf{Binary} & \textbf{Soft} & \textbf{Binary} & \textbf{LogSig} & \textbf{Accuracy $\uparrow$} & \textbf{Uncertainty $\downarrow$}\\ \midrule

Flickr30K Entities & BLIP$_{base}$ & 0.12 & 0.09 & 0.21 & 0.15 & 0.95 & 0.79 & 0.43 & 60.08 & 317 / 14481 \\

(\textsc{test}) & BLIP$_{large}$ & 0.14 & 0.11 & 0.23 & 0.17 & 0.94 & 0.77 & \underline{0.44} & 71.43 & 965 / 14481 \\

 & CLIP$_{gScoreCAM}$ & \textbf{0.21} & \textbf{0.23} & \textbf{0.33} & \textbf{0.34} & \underline{0.92} & \textbf{0.56} & \underline{0.44} & \underline{75.41} & \hspace{1mm} \textbf{0} / 14481 \\
 
 & ALBEF$_{AMC}$ & \underline{0.19} & \underline{0.16} & \underline{0.31} & \underline{0.25} & \textbf{0.86} & \underline{0.60} & \textbf{0.62} & \textbf{87.69} & \underline{10} / 14481 \\ \midrule

RefCOCO+ & BLIP$_{base}$ & 0.10 & 0.06 & 0.18 & 0.10 & 0.98 & 0.75 & \underline{0.39} & \underline{67.90} & 198 / 5726 \\
(\textsc{testA}) & BLIP$_{large}$ & 0.10 & 0.06 & 0.18 & 0.10 & 0.98 & 0.76 & 0.38 & 67.27 & 336 / 5726 \\
& CLIP$_{gScoreCAM}$ & \textbf{0.17} & \textbf{0.15} & \textbf{0.28} & \textbf{0.24} & \underline{0.98} & \underline{0.68} & 0.34 & 63.36 & \hspace{1mm} \textbf{0} / 5726 \\

 & ALBEF$_{AMC}$ & \underline{0.14} & \underline{0.09} & \underline{0.24} & \underline{0.15} & \textbf{0.96} & \textbf{0.66} & \textbf{0.54} & \textbf{78.81} & \underline{14} 
 / 5726 \\ \midrule

RefCOCO+ & BLIP$_{base}$ & 0.11 & 0.06 & 0.20 & 0.11 & 0.98 & 0.86 & \underline{0.31} & 46.96 & 201 / 4889 \\
(\textsc{testB}) & BLIP$_{large}$ & 0.11 & 0.06 & 0.20 & 0.11 & 0.98 & 0.87 & 0.30 & 46.71 & 449 / 4889 \\

& CLIP$_{gScoreCAM}$ & \textbf{0.18} & \textbf{0.19} & \textbf{0.29} & \textbf{0.29} & \underline{0.97} & \underline{0.85} & 0.30 & \underline{49.02} &\hspace{1mm} \textbf{0} / 4889 \\

 & ALBEF$_{AMC}$ & \underline{0.16} & \underline{0.11} & \underline{0.27} & \underline{0.19} & \textbf{0.96} & \textbf{0.76} & \textbf{0.45} & \textbf{64.34} & \underline{16} 
 / 4889 \\ \midrule

SpatialSense & BLIP$_{base}$ & 0.11 & 0.12 & 0.20 & 0.18 & 0.97 & 0.75 & 0.29 & 46.99 & 50 / 1811 \\
(\textsc{triplets}) & BLIP$_{large}$ & 0.12 & 0.12 & 0.21 & 0.19 & 0.97 & \underline{0.73} & \underline{0.31} & \underline{51.68} & 100 / 1811 \\

& CLIP$_{gScoreCAM}$ & \underline{0.16} & \textbf{0.20} & \underline{0.26} & \textbf{0.30} & \underline{0.95} & 0.81 & 0.28 & 50.57 &\hspace{1mm} \textbf{0} / 1811 \\

& ALBEF$_{AMC}$ & \textbf{0.17} & \underline{0.16} & \textbf{0.27} & \underline{0.26} & \textbf{0.93} & \textbf{0.71} & \textbf{0.42} & \textbf{67.64} & \underline{3} / 1811 \\ \midrule

SpatialSense & BLIP$_{base}$ & 0.10 & 0.08 & 0.17 & 0.14 & 0.97 & 0.76 & 0.29 & 39.81 & 42 / 1811 \\
(\textsc{subjects}) & BLIP$_{large}$ & 0.11 & 0.11 & 0.20 & 0.17 & 0.97 & 0.74 & 0.30 & 49.75 & 116 / 1811 \\

& CLIP$_{gScoreCAM}$ & \textbf{0.17} & \textbf{0.23} & \textbf{0.28} & \textbf{0.34} & \underline{0.95} & \textbf{0.69} & \underline{0.31} & \underline{60.07} & \hspace{1mm} \textbf{0} / 1811 \\

& ALBEF$_{AMC}$ & \underline{0.16} & \underline{0.15} & \underline{0.27} & \underline{0.24} & \textbf{0.89} & \underline{0.70} & \textbf{0.52} & \textbf{70.01} & \underline{4} / 1811 \\ \midrule

SpatialSense & BLIP$_{base}$ & 0.11 & 0.07 & 0.19 & 0.12 & 0.92 & \underline{0.69} & \underline{0.42} & 50.02 & 37 / 1811 \\
(\textsc{objects}) & BLIP$_{large}$ & 0.12 & 0.08 & 0.21 & 0.13 & 0.92 & 0.74 & 0.40 & 54.55 & 158 / 1811 \\

& CLIP$_{gScoreCAM}$ & \textbf{0.20} & \textbf{0.21} & \textbf{0.32} & \textbf{0.32} & \underline{0.90} & \textbf{0.65} & 0.41 & \underline{64.93} & \hspace{1mm} \textbf{0} / 1811 \\

& ALBEF$_{AMC}$ & \underline{0.16} & \underline{0.12} & \underline{0.27} & \underline{0.19} & \textbf{0.81} & \textbf{0.65} & \textbf{0.63} & \textbf{78.24} & \underline{3} / 1811 \\ \bottomrule
\end{tabular}
}
\caption{Quantitative results comparison for all settings where numbers are reported as $\mathrm{mean}$ $\mathrm{IoU}$ ($\mathbf{mIoU}$) across each setting. For each setting, the top performance across all models is highlighted as \textbf{bold}, and the second-best with \underline{underline}.}
    \label{tab:benchmarking}
\end{table*}
\section{Method}
Consider $\mathrm{GradCAM}$ activation map  $A_{i,j}$, obtained by passing an image and a corresponding text prompt to the model, and $M_{i, j}$ are pixel locations of the binary ground-truth bounding box mask. 
For computing $\mathrm{IoU}_{Soft}$ and $\mathrm{Dice}_{Soft}$, we've followed the formulas used for semantic segmentation, according to \cite{ssegmetrics}.
We first apply the threshold of 0.5 to each pixel value in $A_{i, j}$
and pass the thresholded binary activation maps
to compute the $\mathrm{IoU}_{Binary}$ and $\mathrm{Dice}_{Binary}$.
\noindent\textbf{Distance Maps.} Given the bounding box coordinates as $(y_{0}, x_{0}, y_{1}, x_{1})$, distance maps $D_{i, j}$ is computed as for all $(i, j)$ where $0 \leq i < H, 0 \leq j < W$:
\[ D_{i,j} = \max\{\max(y_{0}-i, i-y_{1}), 
\max(x_{0}-j, j-x_{1}))\}
\]
\noindent And the weighted penalties matrix $P_{i,j}$ is computed as:
\begin{equation}
  P_{i, j} = [A_{i, j} \cdot (1 - M_{i, j}) \cdot D_{i, j}]
\end{equation}
\noindent\textbf{Weighted Distance Penalty (WDP)} is designed to penalize the \textit{spurious} activations outside the ground-truth bounding box, proportional to their \textit{magnitudes} and \textit{distances} from it, computed and normalized as follows:
\begin{equation}
    \mathrm{WDP} = {\tt Sigmoid}\left({\tt log}\frac{\sum_{i=1}^{H} \sum_{j=1}^{W}P_{i, j}}{\sum_{i=1}^{H} \sum_{j=1}^{W}A_{i, j} + \epsilon}\right)
\end{equation}
\noindent\textbf{Inside/Outside Activations Ratio ($\mathrm{IO}_{ratio}$)} is computed as follows:
\begin{align}
S_{\mathrm{inside}} &= \sum_{i=y_{0}}^{y_{1}-1} \sum_{j=x_{0}}^{x_{1}-1} A_{i,j} \\
S_{\mathrm{outside}} &= \sum_{i=1}^{H} \sum_{j=1}^{W} A_{i,j} - S_{\mathrm{inside}}
\end{align}
\begin{equation}
    \mathrm{IO}_{ratio} = {\tt Sigmoid}\left( {\tt log}\frac{S_{\mathrm{inside}}}{S_{\mathrm{outside}}}\right)
\end{equation}
\noindent\textbf{Pointing Game Uncertainty Analysis:} 
We extract the local maxima ($v$) of the activation map $A_{i, j}$ higher than the activation threshold ($\tau$) of $0.7$ as the set of $V = \left\{v_{1}, v_{2}, ..., v_{n}\right\}$. $V$ is then sorted, followed by an additional NMS step of suppressing maxima that are within Euclidean distance $\delta$~\footnote{$\delta=50$ in our experiments} of larger extremum.
$V_{nms}$ and $C_{nms}$, represent remaining activation values and their corresponding coordinates. 
For every point in $C_{nms}$, we check whether it falls inside the ground-truth bounding box and count the number of cases where top-$k$ equal activations in $V_{nms}$ are \underline{\textbf{not}} all either inside or outside of the bounding box.
\begin{figure*}[!h]
  \centering
  \includegraphics[width=\linewidth]{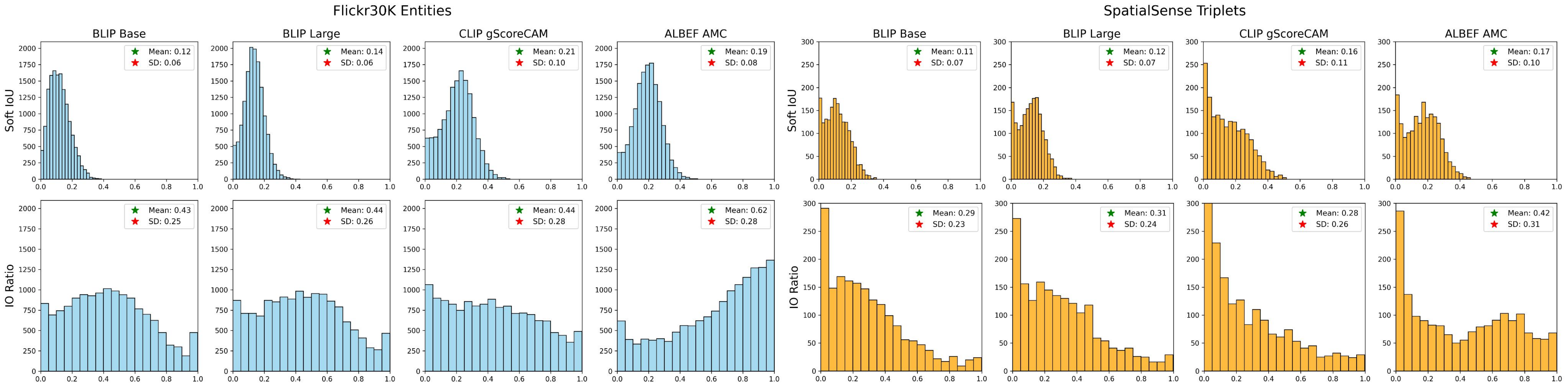}
  \caption{Histogram of $\mathrm{IoU}_{Soft}$ and $\mathrm{IO}_{ratio}$ distributions for \textbf{\textcolor{CornflowerBlue}{ID}} vs. \textbf{\textcolor{BurntOrange}{OOD}}. Note that the histograms are more peaked for in-distribution datasets, as shown in \textbf{\textcolor{CornflowerBlue}{blue}} on the left, and for better-performing models, they are shifted to the right. The out-of-distribution experiments for all models have less peaked, flatter histograms, where shown in \textbf{\textcolor{BurntOrange}{orange}} on the right. Full visualizations can be found in  Appendix~\ref{app:histograms}.}
   \label{fig:mainhistograms}
\end{figure*}
\section{Experiments}
We've conducted experiments to evaluate the grounding ability of four state-of-the-art VLMs ($\mathrm{BLIP}_{base}$, $\mathrm{BLIP}_{large}$, $\mathrm{CLIP}$ $\mathrm{gScoreCAM}$, and the $\mathrm{AMC}$ variation of $\mathrm{ALBEF}$)
on a wide range of grounding tasks, varied by the 
text prompt granularity, and with both in-distribution (ID) and out-of-distribution (OOD) data.
The quantitative results are summarized in Table~\ref{tab:benchmarking}, and sample score distributions are shown in Figure~\ref{fig:mainhistograms}. 
For phrase grounding and referring expression comprehension, we used the test split of Flickr30K Entities~\cite{flickr30ke} and RefCOCO+ testA \& testB \cite{referitgame} datasets that are considered from in-domain distribution for the models.
In order to investigate the out-of-distribution generalizability of these models, we ran the same experiment on the SpatialSense \cite{yang2019spatialsense} dataset test split, 
which has both \textbf{visual} domain shift due to the inclusion of NYU dataset~\cite{silberman2012indoornyu} images with a total of 3,679 unique objects, including small/long-tail concepts, and the \textbf{language} domain shift by the annotation of 17,498 triplets including spatial relations.
The dataset is designed for spatial relationship recognition and is annotated in the triplet \{\textsc{Subject}-\textsc{Relation}-\textsc{Object}\} format with the ground-truth \textsc{Object} and \textsc{Subject} bounding boxes. We ran our experiments in three different settings for \textsc{Triplet}, \textsc{Subject}, and \textsc{Object} grounding. An instance of the SpatialSense dataset is demonstrated in Figure~\ref{fig:ss_main}. Experiment details for all settings can be found in the Appendix~\ref{app:expDetails}. Additional qualitative results can be found in Appendix~\ref{app:sampleQualitativeResultsFull}, where another case of \textbf{Scenario 1} uncertainty is shown in Figure~\ref{fig:refcocoplusB}, obtained by $\mathrm{BLIP}_{base}$ in the top-left sub-figure of the first example.
\begin{figure}[!h]
\centering
\begin{subfigure}{0.31\columnwidth}
\includegraphics[width=\linewidth]{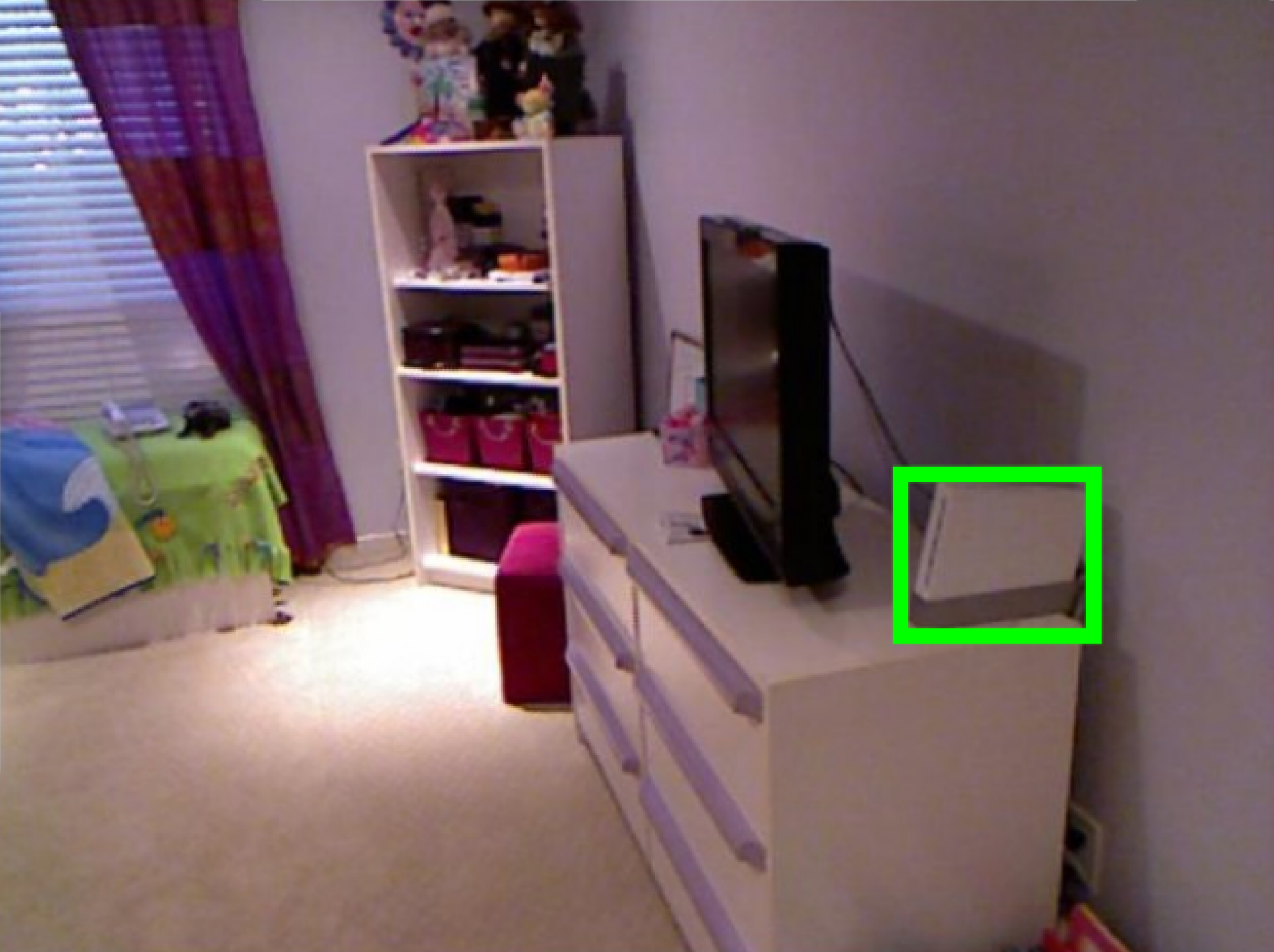}
\caption{Triplet GT-bbox}
\end{subfigure}
\begin{subfigure}{0.31\columnwidth}
\includegraphics[width=\linewidth]{figure/ss_sub.pdf}
\caption{Subject GT-bbox}
\end{subfigure}
\begin{subfigure}{0.31\columnwidth}
\includegraphics[width=\linewidth]{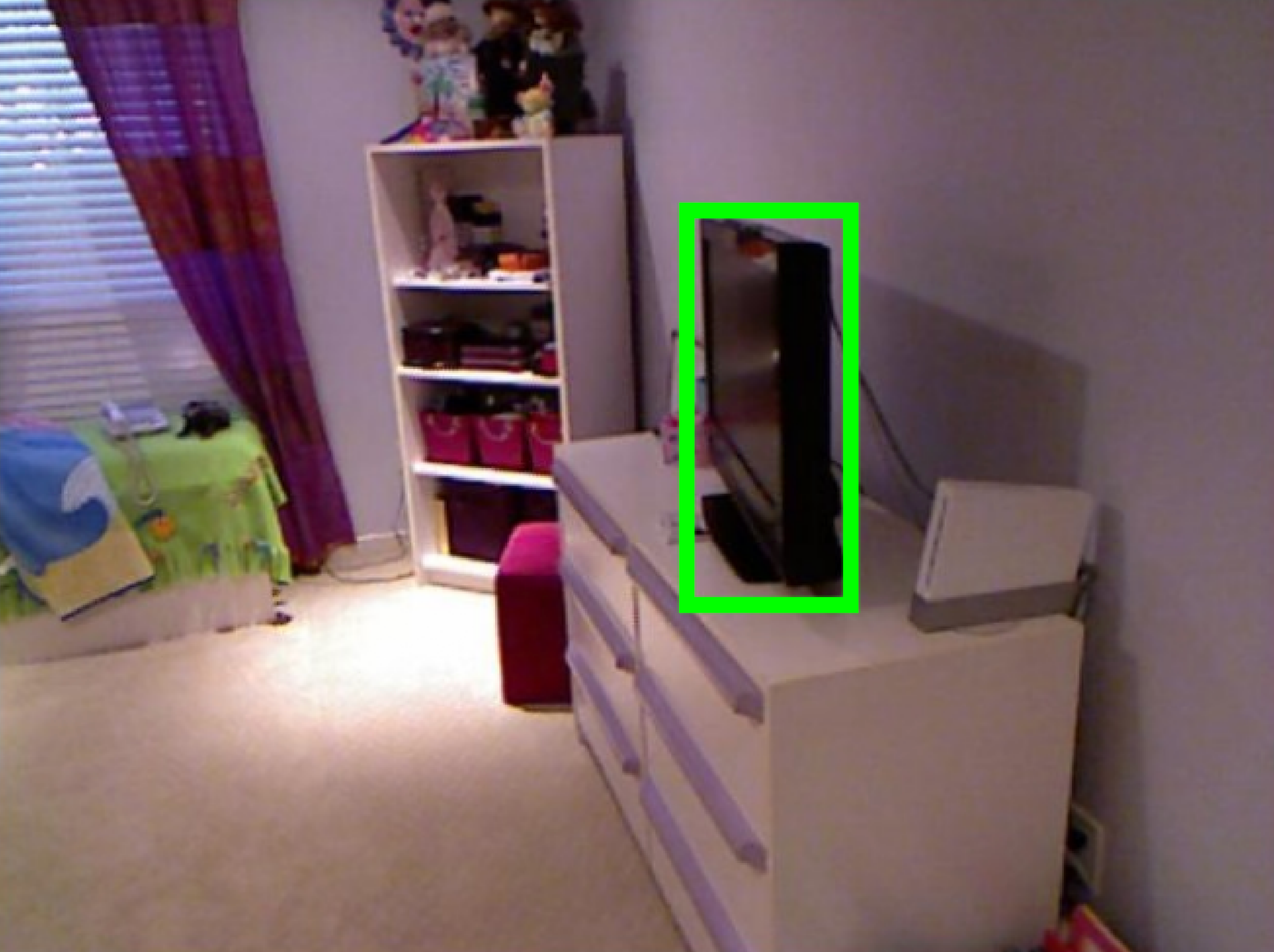}
\caption{Object GT-bbox}
\end{subfigure}
\caption{A sample from SpatialSense NYU bedroom set. We consider the \textit{"wifi router to the right of television"} prompt as \textsc{Triplet}, \textit{"wifi router"} as \textsc{Subject}, and \textit{"television"} as \textsc{Object}.}
\label{fig:ss_main}
\end{figure}
\section{Discussion}
According to Table~\ref{tab:benchmarking}, $\mathrm{ALBEF}_{AMC}$ is the winner, considering the combination of $\mathrm{PG}_{Accuracy}$, $\mathrm{PG}_{Uncertainty}$, and $\mathrm{IO}_{ratio}$ metrics. This highlights the importance of fine-tuning ALBEF bounding box-level supervision, compared to scaling models' size and training set size using often noisy image-text pairs. Regarding the similarity between the activations and ground-truth masks, $\mathrm{CLIP}_{gScoreCAM}$ and $\mathrm{ALBEF}_{AMC}$ are the best and second-best performing models according to IoU and Dice,  while in terms of $\mathrm{WDP}$, it is reversed in most cases. This suggests that CLIP has more spurious GradCAM activations. Furthermore, we believe that due to the prevalent noisiness in $\mathrm{GradCAM}$ activations, the $\mathrm{WDP}_{Soft}$ penalization is much more strict than $\mathrm{WDP}_{Binary}$ variation, making $\mathrm{WDP}_{Binary}$ a more practical metric to use.
$\mathrm{PG}_{Uncertainty}$ is zero in $\mathrm{CLIP}_{gScoreCAM}$, which we believe stems from the difference in the vision backbone and nuances in how the $\mathrm{GradCAM}$ is being computed in ~\cite{chen2022gscorecam}. $\mathrm{ALBEF}_{AMC}$ is the second-best in $\mathrm{PG}_{Uncertainty}$, marginally. In contrast, $\mathrm{BLIP}_{large}$ has shown the highest relative $\mathrm{PG}_{Uncertainty}$ across all the settings.
Our $\mathrm{IO}_{ratio}$ metric has a strong positive correlation with $\mathrm{PG}_{Accuracy}$ while being more strict. This makes it a suitable standalone metric for assessing the model's grounding performance, as it considers both inside \& outside activations, in addition to $\mathrm{PG}_{Accuracy}$. Note that $\mathrm{PG}_{Accuracy}$ is not positively correlated with $\mathrm{PG}_{Uncertainty}$ in all models.
This finding is insightful as it demonstrates that not always the better performing model in terms of $\mathrm{PG}_{Accuracy}$ has the lowest $\mathrm{PG}_{Uncertainty}$. Our OOD experiments also demonstrate the applicability of our metrics to the evaluation of triplet-based spatial understanding grounding~\cite{rajabi2023towards, krishna2017visualgenome}, and shifted visual domains.

\textbf{Model size vs. training data impact.} Apart from the $\mathrm{ALBEF}_{AMC}$ superiority, our experiments show that $\mathrm{BLIP}_{large}$, which has $\sim$ 446M parameters and pre-trained on 129M image-text pairs, \underline{\textit{under-performs}} $\mathrm{BLIP}_{base}$, which has $\sim$ 223M parameters and pre-trained on 14M image-text pairs, in terms of the $\mathrm{PG}_{Uncertainty}$ in all \underline{6} dataset splits, and in \underline{2} dataset splits in terms of $\mathrm{PG}_{Accuracy}$. Considering $\mathrm{CLIP}$, which is pre-trained on $\sim$ 400M image-text pairs, as the highest end, and both $\mathrm{ALBEF}$ \& $\mathrm{BLIP}_{base}$ as the lowest end of the model \& data size spectrum in our experiments, the performance of $\mathrm{ALBEF}_{AMC}$ again highlights the effectiveness of finer-grained fine-tuning.
We suggest running our $\mathrm{IO}_{ratio}$, $\mathrm{PG}_{Uncertainty}$, $\mathrm{WDP}_{Binary}$, and either of $\mathrm{IoU}_{Soft}$ or $\mathrm{Dice}_{Soft}$ metrics, in addition to $\mathrm{PG}_{Accuracy}$, for a thorough grounding evaluation.
\section{Conclusion}
We first demonstrate two scenarios that the Pointing Game (PG) evaluation metric fails to handle properly, and introduce a new set of metrics for evaluation of models' grounding ability that captures finer-grained differences between models. According to our experiments, $\mathrm{ALBEF}_{AMC}$ has shown its superiority quantitatively, compared to the other three models, and demonstrated better performance qualitatively in terms of the sharpness of the activations it predicts inside the ground-truth bounding box, as characterized by $\mathrm{IO}_{ratio}$. The proposed metrics enable a finer-grained evaluation of grounding ability for phrase grounding and referring expression comprehension, and how it varies across in-distribution and out-of-distribution datasets. 

{
    \small
    \bibliographystyle{ieeenat_fullname}
    \bibliography{main}

\begin{thebibliography}{34}
\providecommand{\natexlab}[1]{#1}
\providecommand{\url}[1]{\texttt{#1}}
\expandafter\ifx\csname urlstyle\endcsname\relax
  \providecommand{\doi}[1]{doi: #1}\else
  \providecommand{\doi}{doi: \begingroup \urlstyle{rm}\Url}\fi

\bibitem[Akbari et~al.(2019)Akbari, Karaman, Bhargava, Chen, Vondrick, and Chang]{akbari2019multi}
Hassan Akbari, Svebor Karaman, Surabhi Bhargava, Brian Chen, Carl Vondrick, and Shih-Fu Chang.
\newblock Multi-level multimodal common semantic space for image-phrase grounding.
\newblock In \emph{Proceedings of the IEEE/CVF conference on computer vision and pattern recognition}, pages 12476--12486, 2019.

\bibitem[Arbelle et~al.(2021)Arbelle, Doveh, Alfassy, Shtok, Lev, Schwartz, Kuehne, Levi, Sattigeri, Panda, et~al.]{arbelle2021detector}
Assaf Arbelle, Sivan Doveh, Amit Alfassy, Joseph Shtok, Guy Lev, Eli Schwartz, Hilde Kuehne, Hila~Barak Levi, Prasanna Sattigeri, Rameswar Panda, et~al.
\newblock Detector-free weakly supervised grounding by separation.
\newblock In \emph{Proceedings of the IEEE/CVF International Conference on Computer Vision}, pages 1801--1812, 2021.

\bibitem[Chen et~al.(2022)Chen, Li, Biaz, Bui, and Nguyen]{chen2022gscorecam}
Peijie Chen, Qi Li, Saad Biaz, Trung Bui, and Anh Nguyen.
\newblock gscorecam: What objects is clip looking at?
\newblock In \emph{Proceedings of the Asian Conference on Computer Vision}, pages 1959--1975, 2022.

\bibitem[Datta et~al.(2019)Datta, Sikka, Roy, Ahuja, Parikh, and Divakaran]{datta2019align2ground}
Samyak Datta, Karan Sikka, Anirban Roy, Karuna Ahuja, Devi Parikh, and Ajay Divakaran.
\newblock Align2ground: Weakly supervised phrase grounding guided by image-caption alignment.
\newblock In \emph{Proceedings of the IEEE/CVF international conference on computer vision}, pages 2601--2610, 2019.

\bibitem[Dosovitskiy et~al.(2020)Dosovitskiy, Beyer, Kolesnikov, Weissenborn, Zhai, Unterthiner, Dehghani, Minderer, Heigold, Gelly, et~al.]{dosovitskiy2020imagevit}
Alexey Dosovitskiy, Lucas Beyer, Alexander Kolesnikov, Dirk Weissenborn, Xiaohua Zhai, Thomas Unterthiner, Mostafa Dehghani, Matthias Minderer, Georg Heigold, Sylvain Gelly, et~al.
\newblock An image is worth 16x16 words: Transformers for image recognition at scale.
\newblock \emph{arXiv preprint arXiv:2010.11929}, 2020.

\bibitem[Dou et~al.(2022)Dou, Kamath, Gan, Zhang, Wang, Li, Liu, Liu, LeCun, Peng, et~al.]{dou2022coarsefiber}
Zi-Yi Dou, Aishwarya Kamath, Zhe Gan, Pengchuan Zhang, Jianfeng Wang, Linjie Li, Zicheng Liu, Ce Liu, Yann LeCun, Nanyun Peng, et~al.
\newblock Coarse-to-fine vision-language pre-training with fusion in the backbone.
\newblock \emph{Advances in neural information processing systems}, 35:\penalty0 32942--32956, 2022.

\bibitem[Gupta et~al.(2020)Gupta, Vahdat, Chechik, Yang, Kautz, and Hoiem]{gupta2020contrastive}
Tanmay Gupta, Arash Vahdat, Gal Chechik, Xiaodong Yang, Jan Kautz, and Derek Hoiem.
\newblock Contrastive learning for weakly supervised phrase grounding.
\newblock In \emph{European Conference on Computer Vision}, pages 752--768. Springer, 2020.

\bibitem[Gupta et~al.(2022)Gupta, Kamath, Kembhavi, and Hoiem]{gpv}
Tanmay Gupta, Amita Kamath, Aniruddha Kembhavi, and Derek Hoiem.
\newblock Towards general purpose vision systems: An end-to-end task-agnostic vision-language architecture.
\newblock In \emph{Proceedings of the IEEE/CVF Conference on Computer Vision and Pattern Recognition (CVPR)}, pages 16399--16409, 2022.

\bibitem[Hendricks and Nematzadeh(2021)]{svo}
Lisa~Anne Hendricks and Aida Nematzadeh.
\newblock Probing image-language transformers for verb understanding.
\newblock \emph{arXiv preprint arXiv:2106.09141}, 2021.

\bibitem[Huang et~al.(2023)Huang, Mees, Zeng, and Burgard]{huang2023vlmaps}
Chenguang Huang, Oier Mees, Andy Zeng, and Wolfram Burgard.
\newblock Visual language maps for robot navigation.
\newblock In \emph{2023 IEEE International Conference on Robotics and Automation (ICRA)}, pages 10608--10615. IEEE, 2023.

\bibitem[Kamath et~al.(2021)Kamath, Singh, LeCun, Synnaeve, Misra, and Carion]{kamath2021mdetr}
Aishwarya Kamath, Mannat Singh, Yann LeCun, Gabriel Synnaeve, Ishan Misra, and Nicolas Carion.
\newblock Mdetr-modulated detection for end-to-end multi-modal understanding.
\newblock In \emph{Proceedings of the IEEE/CVF International Conference on Computer Vision}, pages 1780--1790, 2021.

\bibitem[Kazemzadeh et~al.(2014)Kazemzadeh, Ordonez, Matten, and Berg]{referitgame}
Sahar Kazemzadeh, Vicente Ordonez, Mark Matten, and Tamara Berg.
\newblock Referitgame: Referring to objects in photographs of natural scenes.
\newblock In \emph{Proceedings of the 2014 conference on empirical methods in natural language processing (EMNLP)}, pages 787--798, 2014.

\bibitem[Krishna et~al.(2017)Krishna, Zhu, Groth, Johnson, Hata, Kravitz, Chen, Kalantidis, Li, Shamma, et~al.]{krishna2017visualgenome}
Ranjay Krishna, Yuke Zhu, Oliver Groth, Justin Johnson, Kenji Hata, Joshua Kravitz, Stephanie Chen, Yannis Kalantidis, Li-Jia Li, David~A Shamma, et~al.
\newblock Visual genome: Connecting language and vision using crowdsourced dense image annotations.
\newblock \emph{International journal of computer vision}, 123:\penalty0 32--73, 2017.

\bibitem[Li et~al.(2023)Li, Li, Le, Wang, Savarese, and Hoi]{li-etal-2023-lavis}
Dongxu Li, Junnan Li, Hung Le, Guangsen Wang, Silvio Savarese, and Steven~C.H. Hoi.
\newblock {LAVIS}: A one-stop library for language-vision intelligence.
\newblock In \emph{Proceedings of the 61st Annual Meeting of the Association for Computational Linguistics (Volume 3: System Demonstrations)}, pages 31--41, Toronto, Canada, 2023. Association for Computational Linguistics.

\bibitem[Li et~al.(2021)Li, Selvaraju, Gotmare, Joty, Xiong, and Hoi]{albef}
Junnan Li, Ramprasaath Selvaraju, Akhilesh Gotmare, Shafiq Joty, Caiming Xiong, and Steven Chu~Hong Hoi.
\newblock Align before fuse: Vision and language representation learning with momentum distillation.
\newblock \emph{Advances in neural information processing systems}, 34:\penalty0 9694--9705, 2021.

\bibitem[Li et~al.(2022)Li, Li, Xiong, and Hoi]{blip}
Junnan Li, Dongxu Li, Caiming Xiong, and Steven Hoi.
\newblock Blip: Bootstrapping language-image pre-training for unified vision-language understanding and generation.
\newblock In \emph{International Conference on Machine Learning}, pages 12888--12900. PMLR, 2022.

\bibitem[Lin et~al.(2014)Lin, Maire, Belongie, Hays, Perona, Ramanan, Doll{\'a}r, and Zitnick]{mscoco}
Tsung-Yi Lin, Michael Maire, Serge Belongie, James Hays, Pietro Perona, Deva Ramanan, Piotr Doll{\'a}r, and C~Lawrence Zitnick.
\newblock Microsoft coco: Common objects in context.
\newblock In \emph{Computer Vision--ECCV 2014: 13th European Conference, Zurich, Switzerland, September 6-12, 2014, Proceedings, Part V 13}, pages 740--755. Springer, 2014.

\bibitem[Liu et~al.(2023)Liu, Li, Wu, and Lee]{liu2023llava}
Haotian Liu, Chunyuan Li, Qingyang Wu, and Yong~Jae Lee.
\newblock Visual instruction tuning.
\newblock In \emph{NeurIPS}, 2023.

\bibitem[Monteux(2019)]{ssegmetrics}
Angelo Monteux.
\newblock Metrics for semantic segmentation, 2019.

\bibitem[Peng et~al.(2023)Peng, Wang, Dong, Hao, Huang, Ma, and Wei]{peng2023kosmos2}
Zhiliang Peng, Wenhui Wang, Li Dong, Yaru Hao, Shaohan Huang, Shuming Ma, and Furu Wei.
\newblock Kosmos-2: Grounding multimodal large language models to the world.
\newblock \emph{arXiv preprint arXiv:2306.14824}, 2023.

\bibitem[Plummer et~al.(2015)Plummer, Wang, Cervantes, Caicedo, Hockenmaier, and Lazebnik]{flickr30ke}
Bryan~A. Plummer, Liwei Wang, Chris~M. Cervantes, Juan~C. Caicedo, Julia Hockenmaier, and Svetlana Lazebnik.
\newblock Flickr30k entities: Collecting region-to-phrase correspondences for richer image-to-sentence models.
\newblock In \emph{Proceedings of the IEEE International Conference on Computer Vision (ICCV)}, 2015.

\bibitem[Qu et~al.(2022)Qu, Wu, Liu, Gong, Liang, Russakovsky, Zhao, and Wei]{qu2022siri}
Mengxue Qu, Yu Wu, Wu Liu, Qiqi Gong, Xiaodan Liang, Olga Russakovsky, Yao Zhao, and Yunchao Wei.
\newblock Siri: A simple selective retraining mechanism for transformer-based visual grounding.
\newblock In \emph{European Conference on Computer Vision}, pages 546--562. Springer, 2022.

\bibitem[Radford et~al.(2021)Radford, Kim, Hallacy, Ramesh, Goh, Agarwal, Sastry, Askell, Mishkin, Clark, et~al.]{clip}
Alec Radford, Jong~Wook Kim, Chris Hallacy, Aditya Ramesh, Gabriel Goh, Sandhini Agarwal, Girish Sastry, Amanda Askell, Pamela Mishkin, Jack Clark, et~al.
\newblock Learning transferable visual models from natural language supervision.
\newblock In \emph{International Conference on Machine Learning}, pages 8748--8763. PMLR, 2021.

\bibitem[Rajabi and Kosecka(2023)]{rajabi2023towards}
Navid Rajabi and Jana Kosecka.
\newblock Towards grounded visual spatial reasoning in multi-modal vision language models.
\newblock \emph{arXiv preprint arXiv:2308.09778}, 2023.

\bibitem[Selvaraju et~al.(2017)Selvaraju, Cogswell, Das, Vedantam, Parikh, and Batra]{GradCam}
Ramprasaath~R Selvaraju, Michael Cogswell, Abhishek Das, Ramakrishna Vedantam, Devi Parikh, and Dhruv Batra.
\newblock Grad-cam: Visual explanations from deep networks via gradient-based localization.
\newblock In \emph{Proceedings of the IEEE international conference on computer vision}, pages 618--626, 2017.

\bibitem[Silberman et~al.(2012)Silberman, Hoiem, Kohli, and Fergus]{silberman2012indoornyu}
Nathan Silberman, Derek Hoiem, Pushmeet Kohli, and Rob Fergus.
\newblock Indoor segmentation and support inference from rgbd images.
\newblock In \emph{Computer Vision--ECCV 2012: 12th European Conference on Computer Vision, Florence, Italy, October 7-13, 2012, Proceedings, Part V 12}, pages 746--760. Springer, 2012.

\bibitem[Subramanian et~al.(2022)Subramanian, Merrill, Darrell, Gardner, Singh, and Rohrbach]{reclip}
Sanjay Subramanian, Will Merrill, Trevor Darrell, Matt Gardner, Sameer Singh, and Anna Rohrbach.
\newblock Reclip: A strong zero-shot baseline for referring expression comprehension.
\newblock \emph{arXiv preprint arXiv:2204.05991}, 2022.

\bibitem[Thrush et~al.(2022)Thrush, Jiang, Bartolo, Singh, Williams, Kiela, and Ross]{winoground}
Tristan Thrush, Ryan Jiang, Max Bartolo, Amanpreet Singh, Adina Williams, Douwe Kiela, and Candace Ross.
\newblock Winoground: Probing vision and language models for visio-linguistic compositionality.
\newblock In \emph{Proceedings of the IEEE/CVF Conference on Computer Vision and Pattern Recognition}, pages 5238--5248, 2022.

\bibitem[Wang et~al.(2021)Wang, Huang, Li, Xu, Yang, and Yu]{wang2021improving}
Liwei Wang, Jing Huang, Yin Li, Kun Xu, Zhengyuan Yang, and Dong Yu.
\newblock Improving weakly supervised visual grounding by contrastive knowledge distillation.
\newblock In \emph{Proceedings of the IEEE/CVF conference on computer vision and pattern recognition}, pages 14090--14100, 2021.

\bibitem[Yang et~al.(2019)Yang, Russakovsky, and Deng]{yang2019spatialsense}
Kaiyu Yang, Olga Russakovsky, and Jia Deng.
\newblock Spatialsense: An adversarially crowdsourced benchmark for spatial relation recognition.
\newblock In \emph{International Conference on Computer Vision (ICCV)}, 2019.

\bibitem[Yang et~al.(2023)Yang, Kafle, Dernoncourt, and Ordonez]{amc}
Ziyan Yang, Kushal Kafle, Franck Dernoncourt, and Vicente Ordonez.
\newblock Improving visual grounding by encouraging consistent gradient-based explanations.
\newblock In \emph{Proceedings of the IEEE/CVF Conference on Computer Vision and Pattern Recognition}, pages 19165--19174, 2023.

\bibitem[Young et~al.(2014)Young, Lai, Hodosh, and Hockenmaier]{young2014imageflickr}
Peter Young, Alice Lai, Micah Hodosh, and Julia Hockenmaier.
\newblock From image descriptions to visual denotations: New similarity metrics for semantic inference over event descriptions.
\newblock \emph{Transactions of the Association for Computational Linguistics}, 2:\penalty0 67--78, 2014.

\bibitem[Yuksekgonul et~al.(2022)Yuksekgonul, Bianchi, Kalluri, Jurafsky, and Zou]{bag-of-words}
Mert Yuksekgonul, Federico Bianchi, Pratyusha Kalluri, Dan Jurafsky, and James Zou.
\newblock When and why vision-language models behave like bag-of-words models, and what to do about it?
\newblock \emph{arXiv preprint arXiv:2210.01936}, 2022.

\bibitem[Zhang et~al.(2018)Zhang, Bargal, Lin, Brandt, Shen, and Sclaroff]{pointinggameinit}
Jianming Zhang, Sarah~Adel Bargal, Zhe Lin, Jonathan Brandt, Xiaohui Shen, and Stan Sclaroff.
\newblock Top-down neural attention by excitation backprop.
\newblock \emph{International Journal of Computer Vision}, 126\penalty0 (10):\penalty0 1084--1102, 2018.

\end{thebibliography}
}



\clearpage
\setcounter{page}{1}
\maketitlesupplementary
\section{Experiments Details}
\label{app:expDetails}

For the \textbf{\textit{phrase grounding}} experiments, we used the test split of Flickr30K Entities~\cite{flickr30ke} dataset, specifically, the \textit{merged} version which has also been used by MDETR~\cite{kamath2021mdetr} and SiRi~\cite{qu2022siri}. In terms of the pre-trained checkpoints, we used \texttt{blip-image-text-matching} checkpoint for BLIP (both variations of \texttt{base} and \texttt{large}), through the LAVIS~\cite{li-etal-2023-lavis} codebase. For $\mathrm{ALBEF}_{AMC}$, we used the \texttt{best-flickr} checkpoint released by $\mathrm{AMC}$~\cite{amc} where sample qualitative results are shown in Figure~\ref{fig:introSenario1} \& ~\ref{fig:2ndFigScenario2}. For all $\mathrm{CLIP}_{gScoreCAM}$ experiments, we used the \textsc{ResNet50$\times$16} variation of CLIP with the average pooling of top \texttt{300-channel} activation maps, as the best-performing setting reported by~\cite{chen2022gscorecam}.

For the \textbf{\textit{referring expressions comprehension}} experiments, we used the same checkpoints as \textit{phrase grounding} ones for both $\mathrm{BLIP}$ experiments, but used the \texttt{best-refcoco} checkpoint released by~\cite{amc}, where sample qualitative results are shown in Figure~\ref{fig:refcocoplusA} \&~\ref{fig:refcocoplusB}. Also, an additional example of PG Uncertainty (\textbf{Scenario 1}) is demonstrated in the first example of Figure~\ref{fig:refcocoplusB}, in the top-left $\mathrm{GradCAM}$ activations of the first obtained by running $\mathrm{BLIP}_{base}$. Again, there are two top activations with a value of \texttt{1.0} each, while one of them falls inside and the other one outside of the ground-truth bounding box. In these cases, PG becomes indecisive in picking the maximum to compute the accuracy, as this part of the evaluation becomes stochastic and directly affects the evaluation reliability.

For the \textbf{\textit{Out-of-Ditribution (OOD)}} experiments, we ran the models on the SpatialSense~\cite{yang2019spatialsense} dataset. We consider the Spatial Sense dataset OOD because it has a domain shift in both visual side (due to including new images from Flickr~\cite{young2014imageflickr} and NYU~\cite{silberman2012indoornyu}, instead of widely-used pre-training/fine-tuning datasets like MSCOCO~\cite{mscoco} and Visual Genome~\cite{krishna2017visualgenome}), and language side, by including spatial clauses and small/long-tail objects for increasing the detection/grounding difficulty. To be more specific, we ran this experiment on the instances with \texttt{True} labels in the test set, with a total number of 1811 instances in three settings. The first setting grounds the entire triplet \{\textsc{Subject}-\textsc{Relation}-\textsc{Object}\} and we consider the \textsc{Subject} ground-truth bounding box as the correct bounding box supervision for the entire triplet, since according to their convention, the \textsc{Subject} acts as the target and \textsc{Object} as reference. Also, we have conducted two separate experiments for individual object grounding for \textsc{Subjects} and \textsc{Objects} using their corresponding ground-truth bounding boxes each.
In addition to confirming the high degree of difficulty we have seen during the Spatial Sense experiments, in which we have also shown a sample qualitative example in Figure ~\ref{fig:spatialSenseExample1}, we've noticed that in some of the instances, grounding the \textsc{Subjects} is more complicated due to the underlying ambiguities/distractions when grounding the \textsc{subject} as a standalone phrase. This happens because of the dependency on the \textsc{relation} and \textsc{object} as the context required for disambiguation, which is shown in Figure~\ref{fig:spatialSenseExample2} as an example. Therefore, the \textsc{triplet} and \textsc{object} grounding settings results seem more reliable in general, but since we are comparing different models on exactly the same footing, our \textsc{subject} grounding results should also be considered fair among all the models.

\section{Sample Qualitative Results}
\label{app:sampleQualitativeResultsFull}

\begin{figure*}[!t]
  \centering
  \includegraphics[width=1.\linewidth]{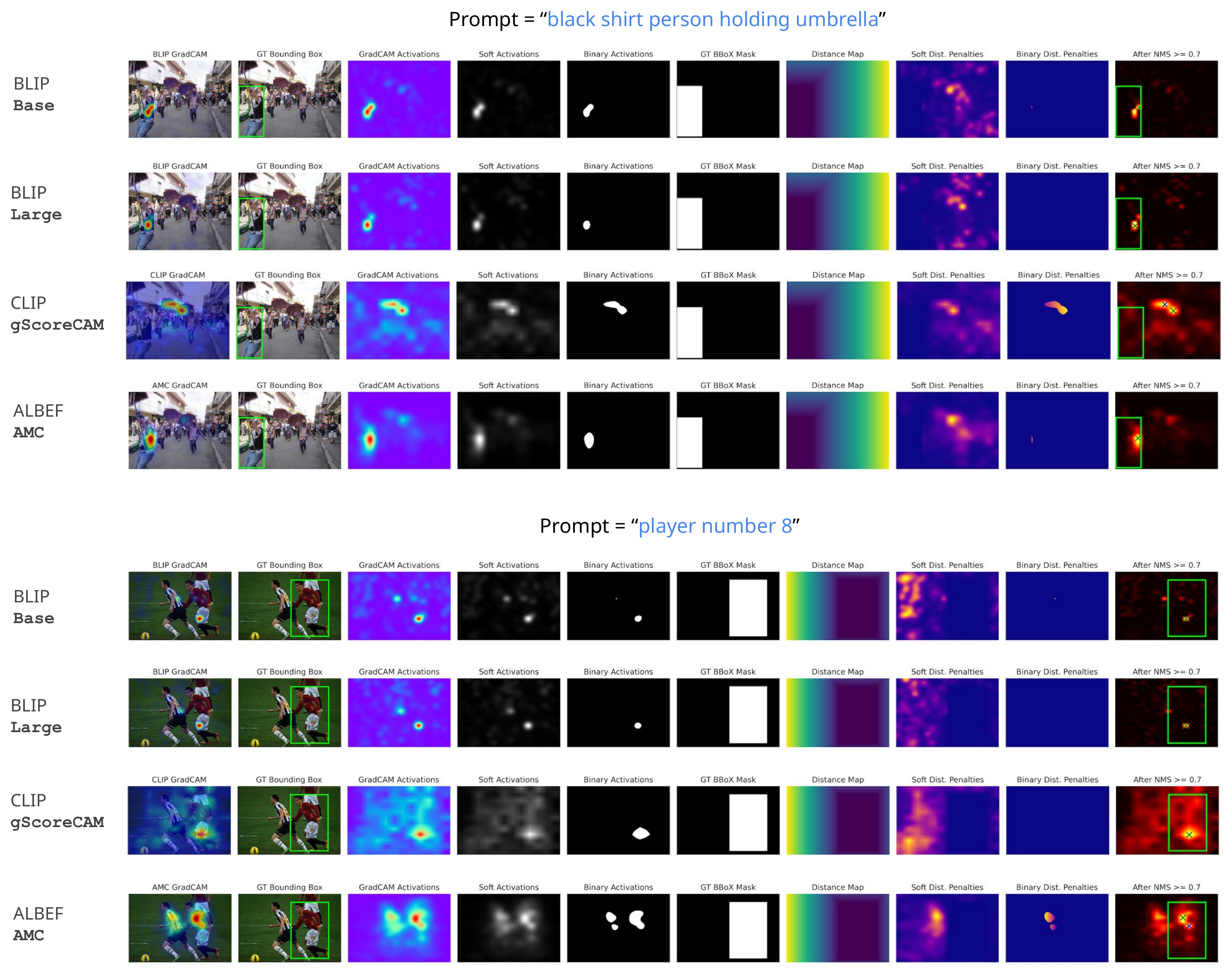}
   \caption{\textbf{RefCOCO+ (testA)} - two sample qualitative results.}
   \label{fig:refcocoplusA}
\end{figure*}

\begin{figure*}[!t]
  \centering
  \includegraphics[width=1.\linewidth]{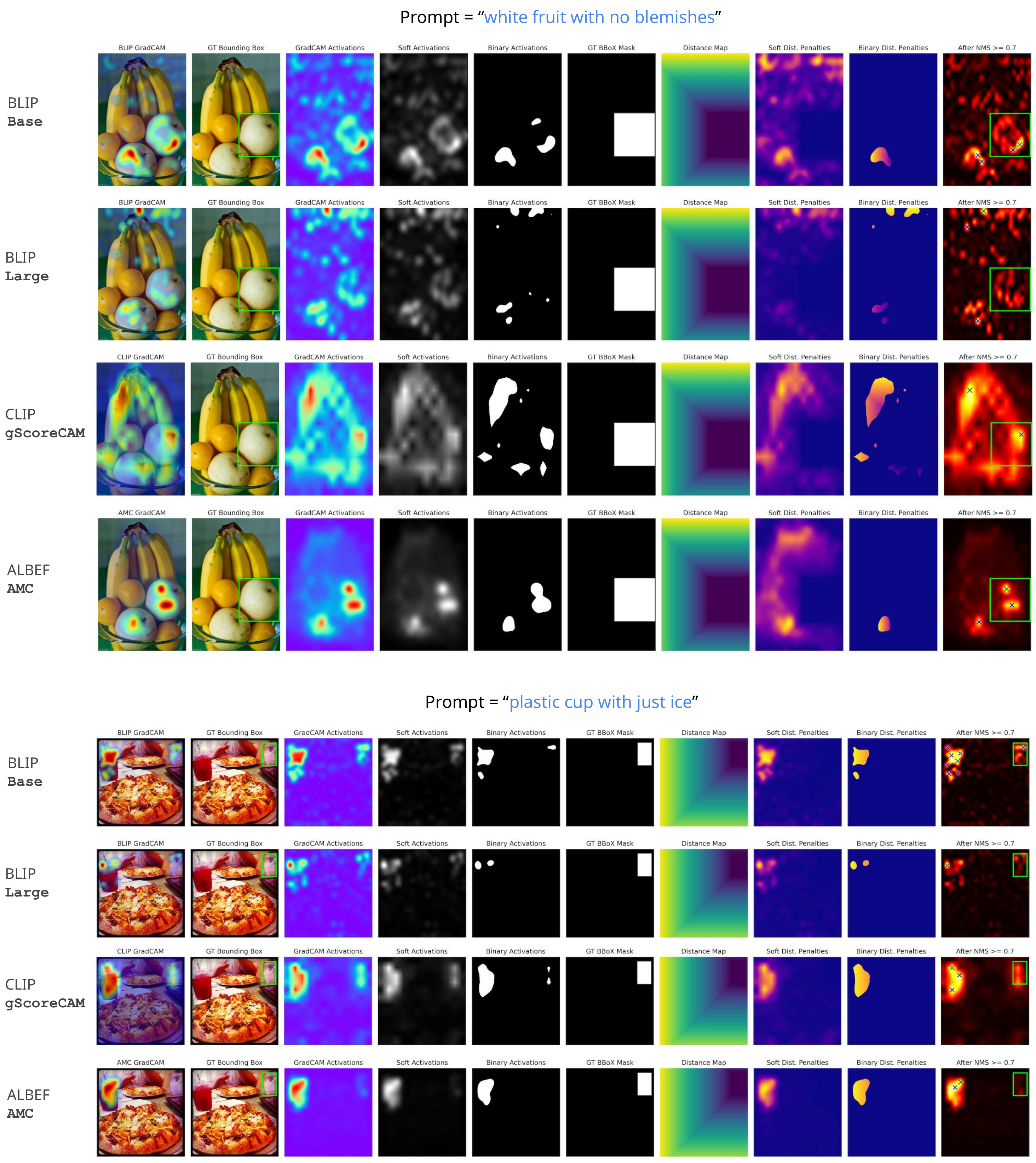}
   \caption{\textbf{RefCOCO+ (testB)} - two sample qualitative results.}
   \label{fig:refcocoplusB}
\end{figure*}

\begin{figure*}[!t]
  \centering
  \includegraphics[width=1.\linewidth]{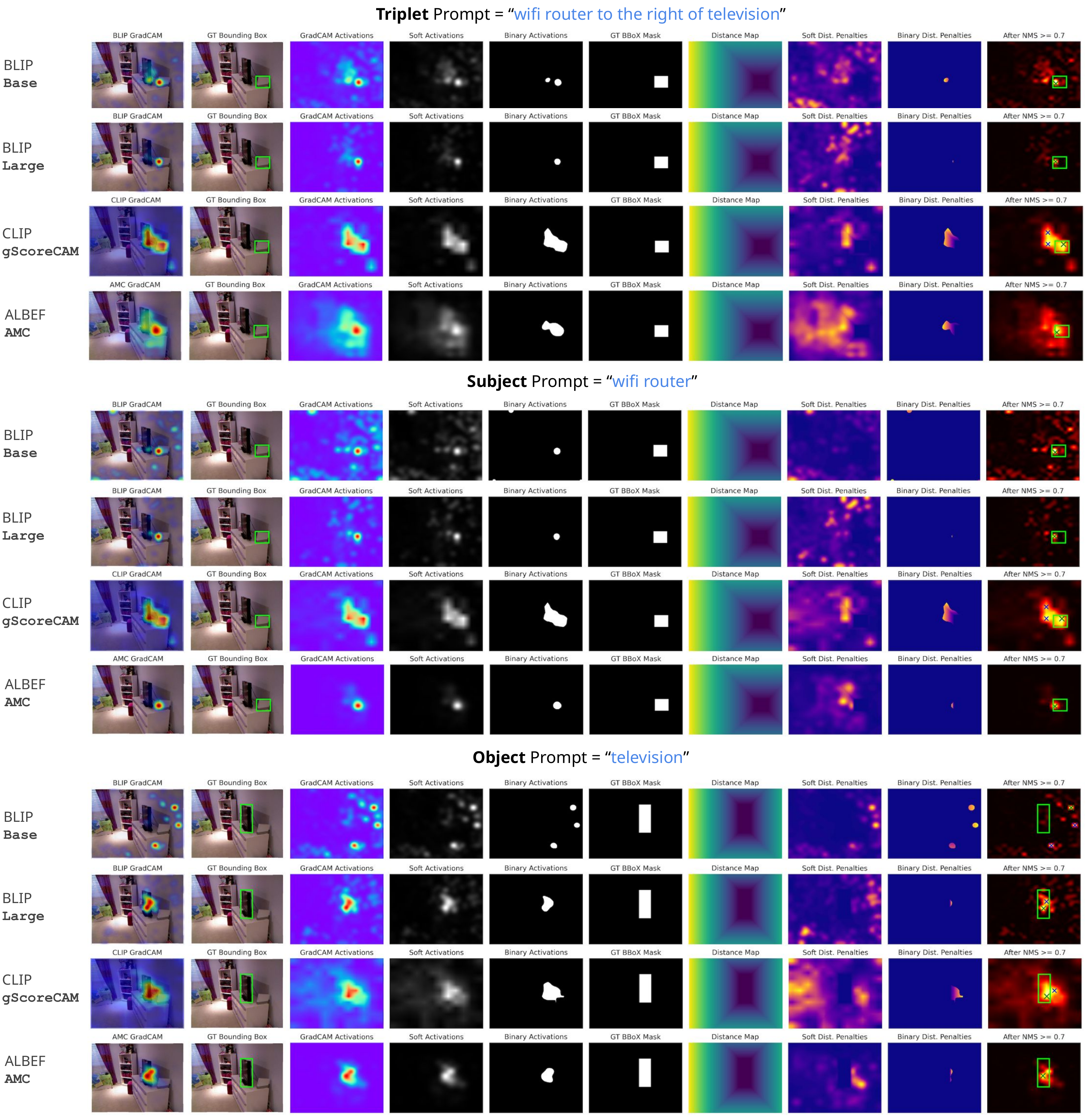}
   \caption{\textbf{Spatial Sense} sample qualitative results \textit{\textbf{without}} the ambiguity for \textsc{Subject} grounding.}
   \label{fig:spatialSenseExample1}
\end{figure*}

\begin{figure*}[!t]
  \centering
  \includegraphics[width=1.\linewidth]{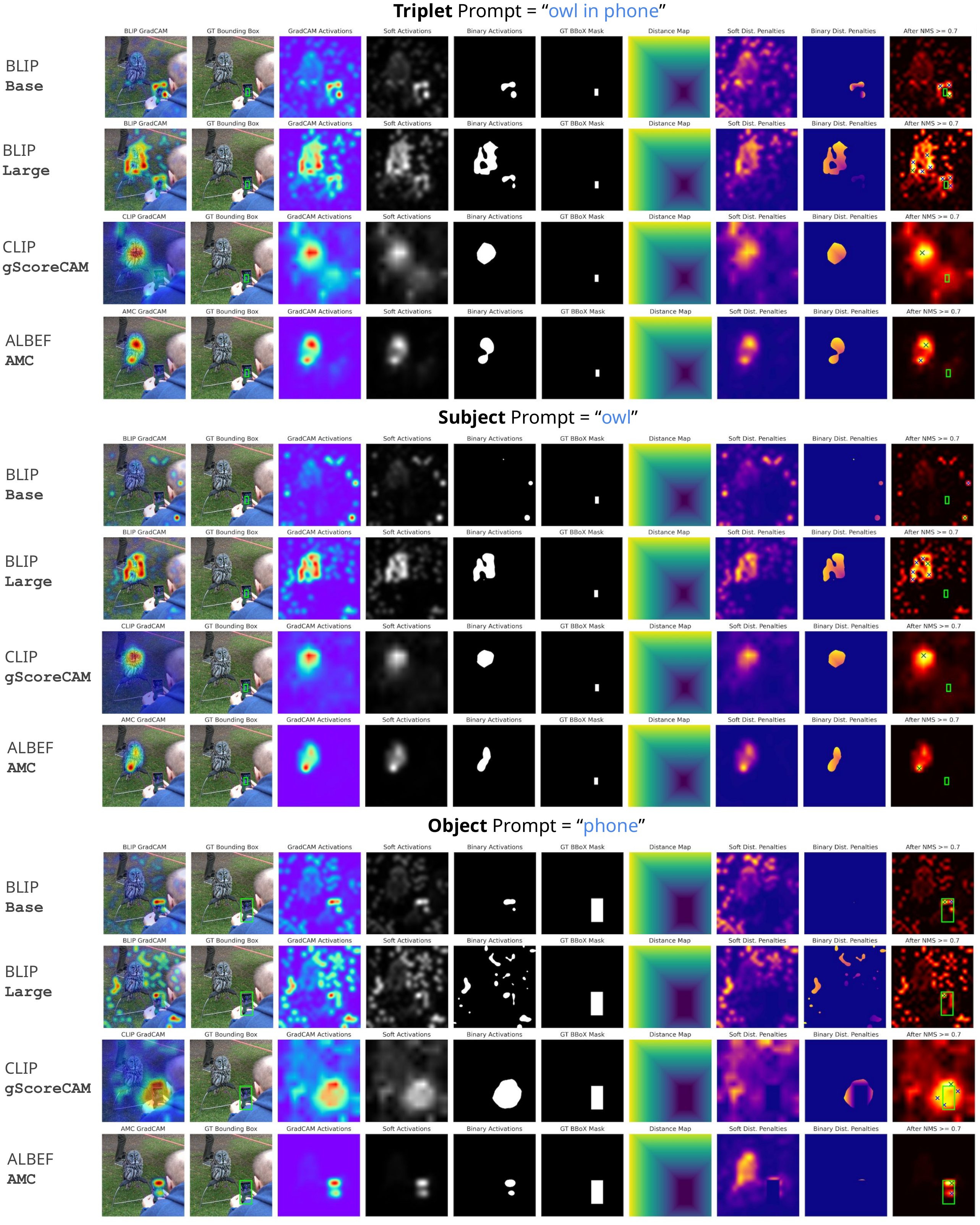}
   \caption{\textbf{Spatial Sense} sample qualitative results \textbf{\textit{with}} ambiguity for \textsc{Subject} grounding, due to the underlying distraction.}
   \label{fig:spatialSenseExample2}
\end{figure*}

Flickr30K Entities examples are provided in Figure~\ref{fig:introSenario1} \&~\ref{fig:2ndFigScenario2}.\\
RefCOCO+ \textbf{testA} example is provided in Figure~\ref{fig:refcocoplusA}.\\
RefCOCO+ \textbf{testB} example is provided in Figure~\ref{fig:refcocoplusB}.\\
SpatialSense \textbf{first} example is provided in Figure~\ref{fig:spatialSenseExample1}.\\
SpatialSense \textbf{second} example is provided in Figure~\ref{fig:spatialSenseExample2}.

\section{Histograms of Score Distributions}
\label{app:histograms}
\begin{figure*}[!t]
  \centering
    \includegraphics[width=1.\linewidth]{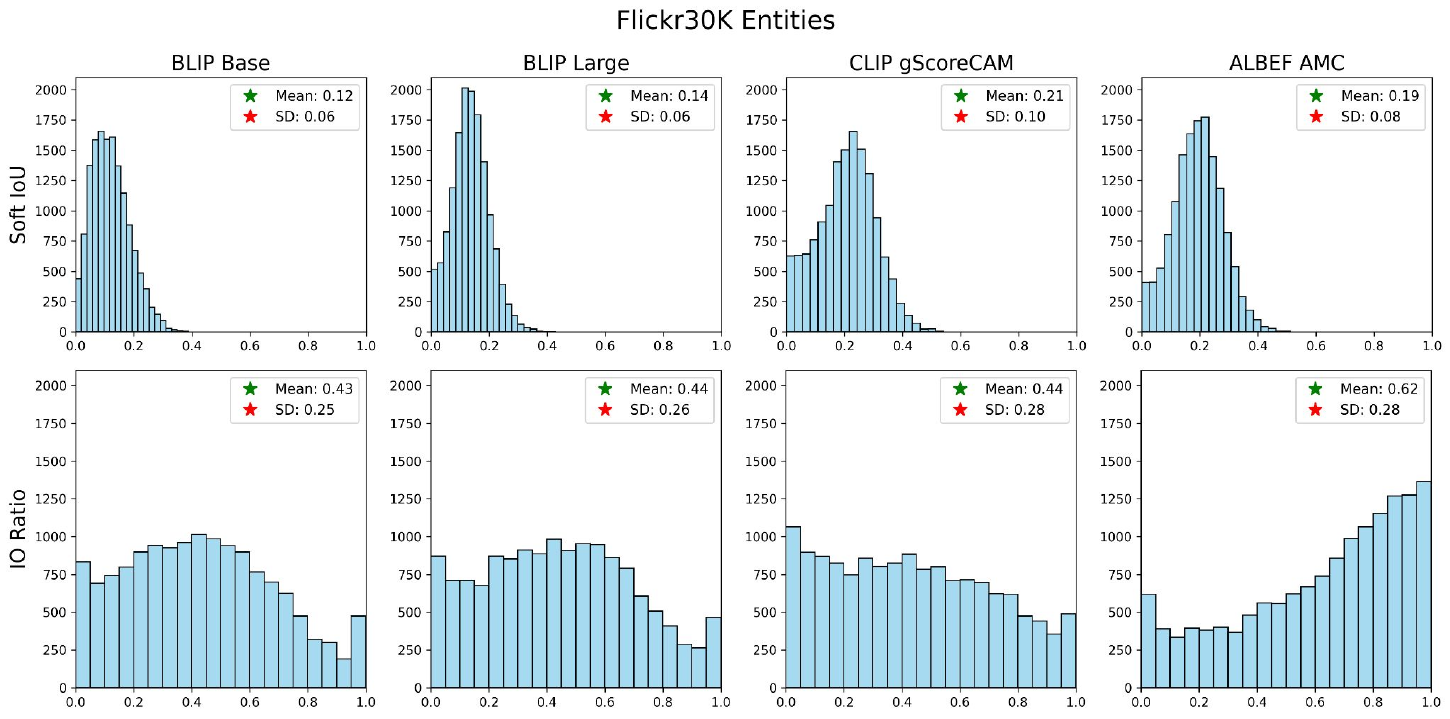}
   \caption{Flickr30K Entities \textbf{test}.}
   \label{fig:hist_viz_fk30_final}
\end{figure*}

\begin{figure*}[!t]
  \centering
  \includegraphics[width=1.\linewidth]{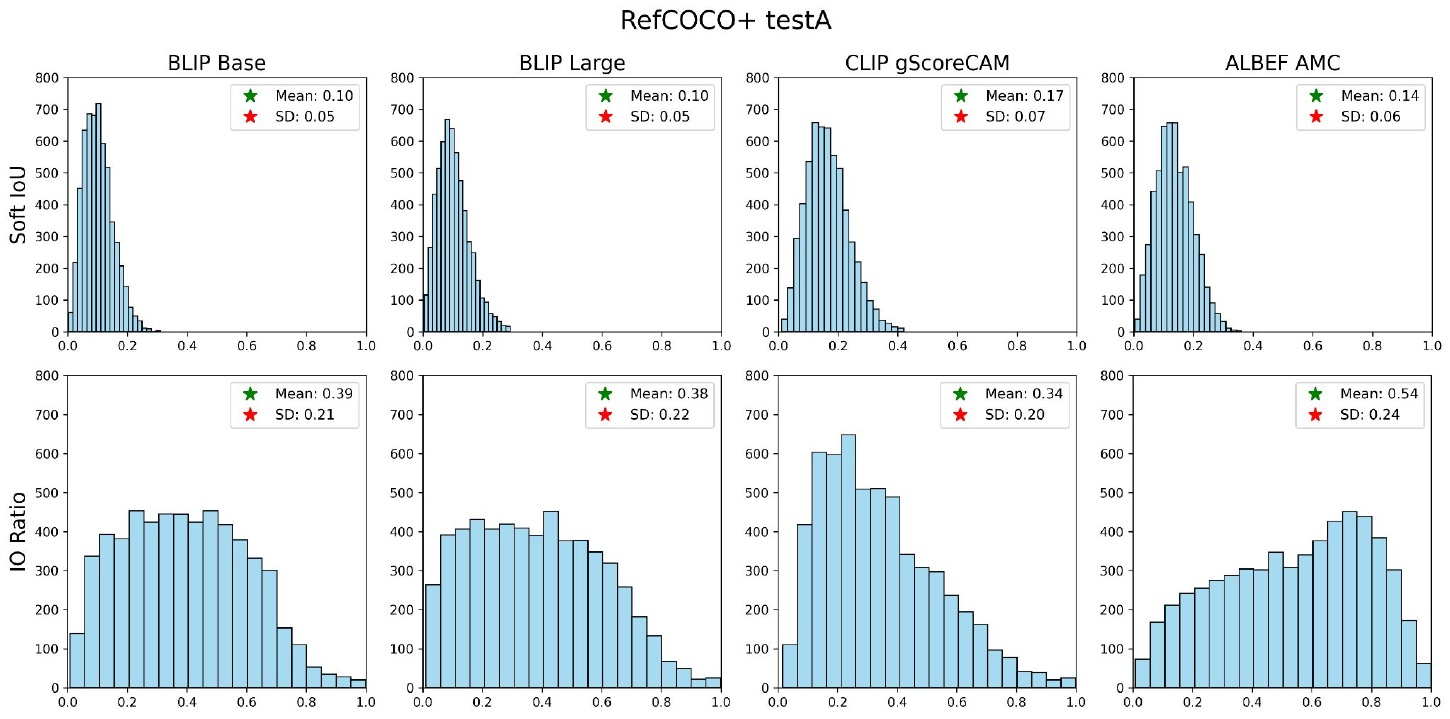}
   \caption{RefCOCO+ \textbf{testA}.}
   \label{fig:refCOCOPlus_testA_hist_viz}
\end{figure*}

\begin{figure*}[!t]
  \centering
  \includegraphics[width=1.\linewidth]{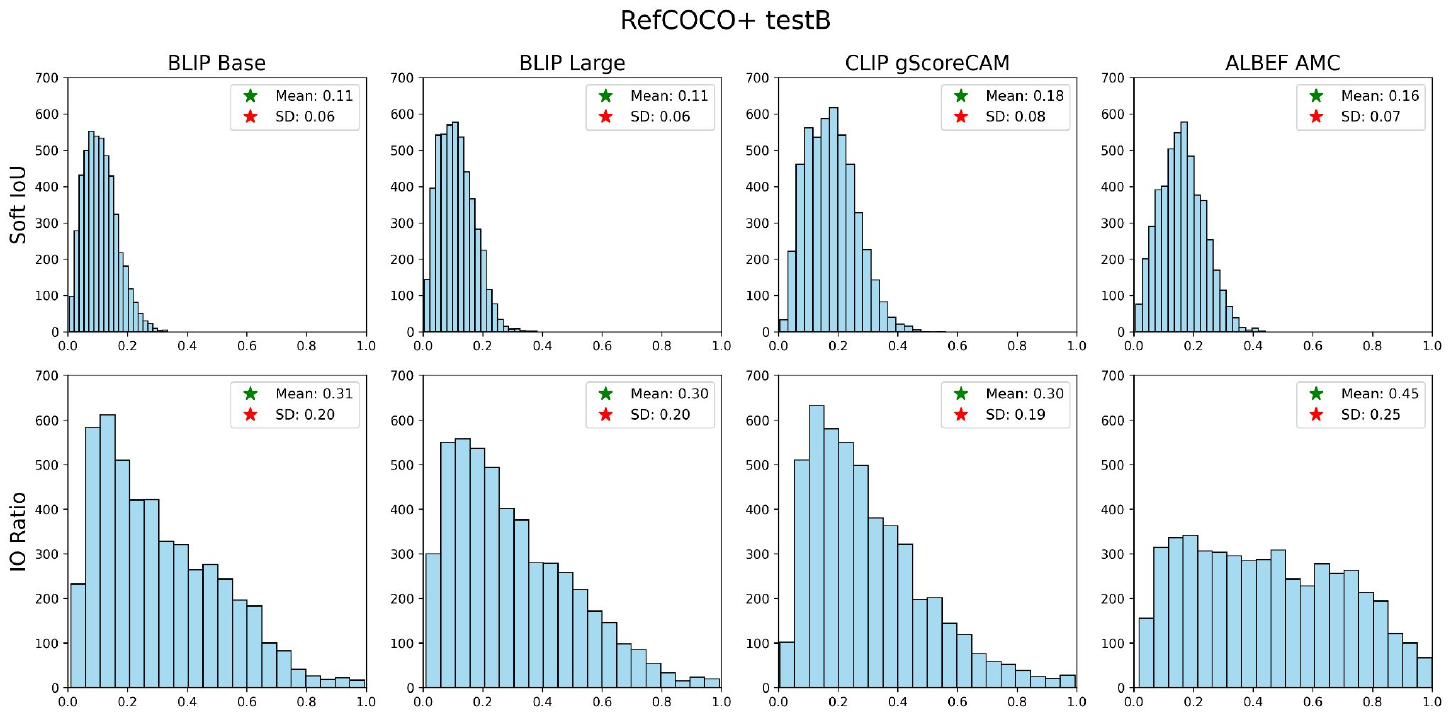}
   \caption{RefCOCO+ \textbf{testB}.}
   \label{fig:refCOCOPlus_testB_hist_viz}
\end{figure*}

\begin{figure*}[!t]
  \centering
  \includegraphics[width=1.\linewidth]{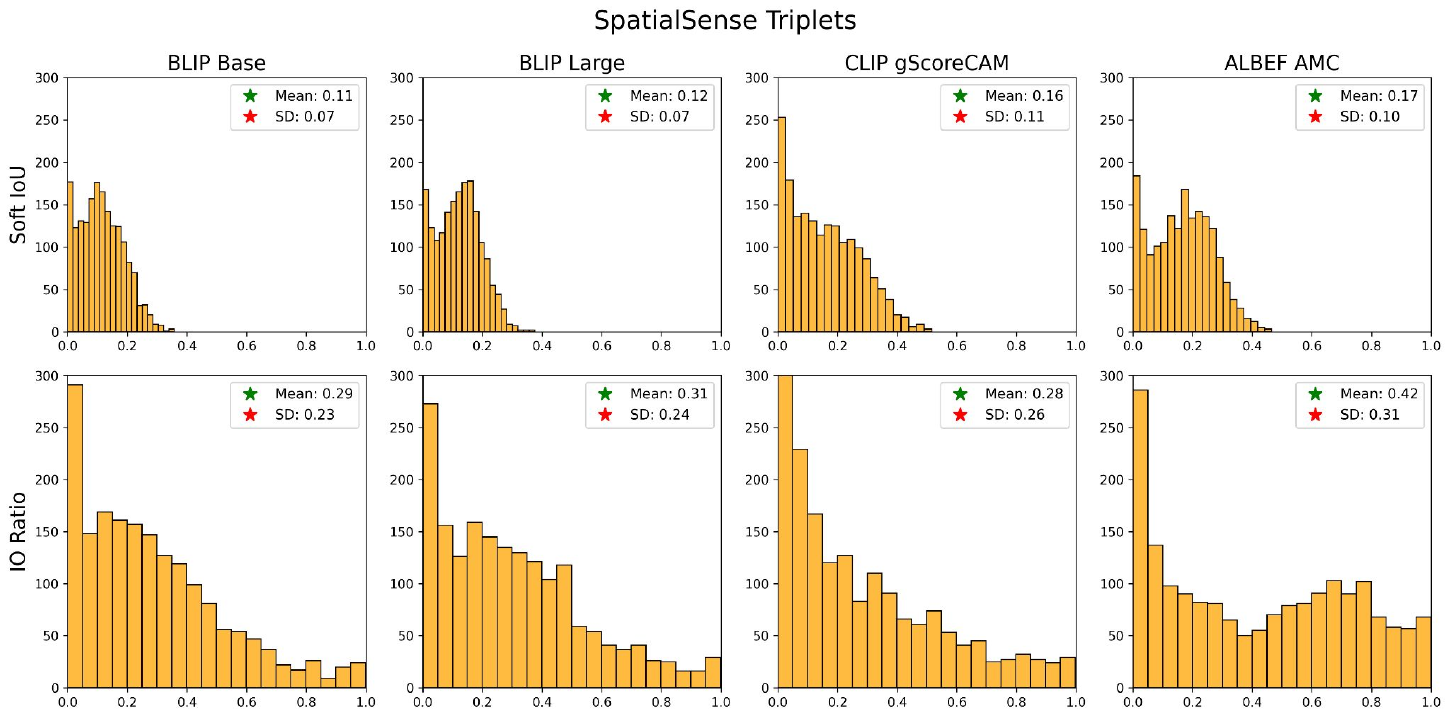}
   \caption{SpatialSense \textbf{triplets}.}
   \label{fig:spatialsense_triplets_hist_viz}
\end{figure*}

\begin{figure*}[!t]
  \centering
  \includegraphics[width=1.\linewidth]{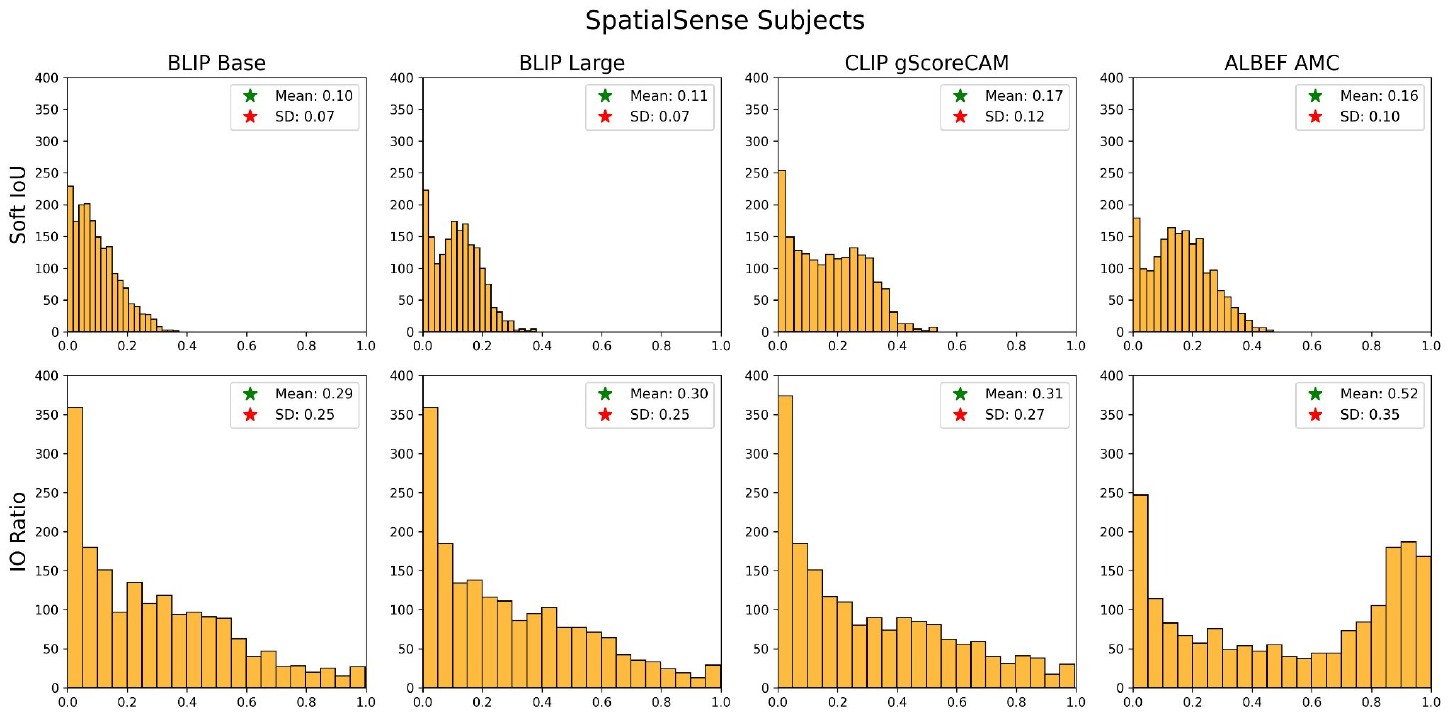}
   \caption{SpatialSense \textbf{subjects}.}
   \label{fig:spatialsense_subjects_hist_viz}
\end{figure*}

\begin{figure*}[!t]
  \centering
  \includegraphics[width=1.\linewidth]{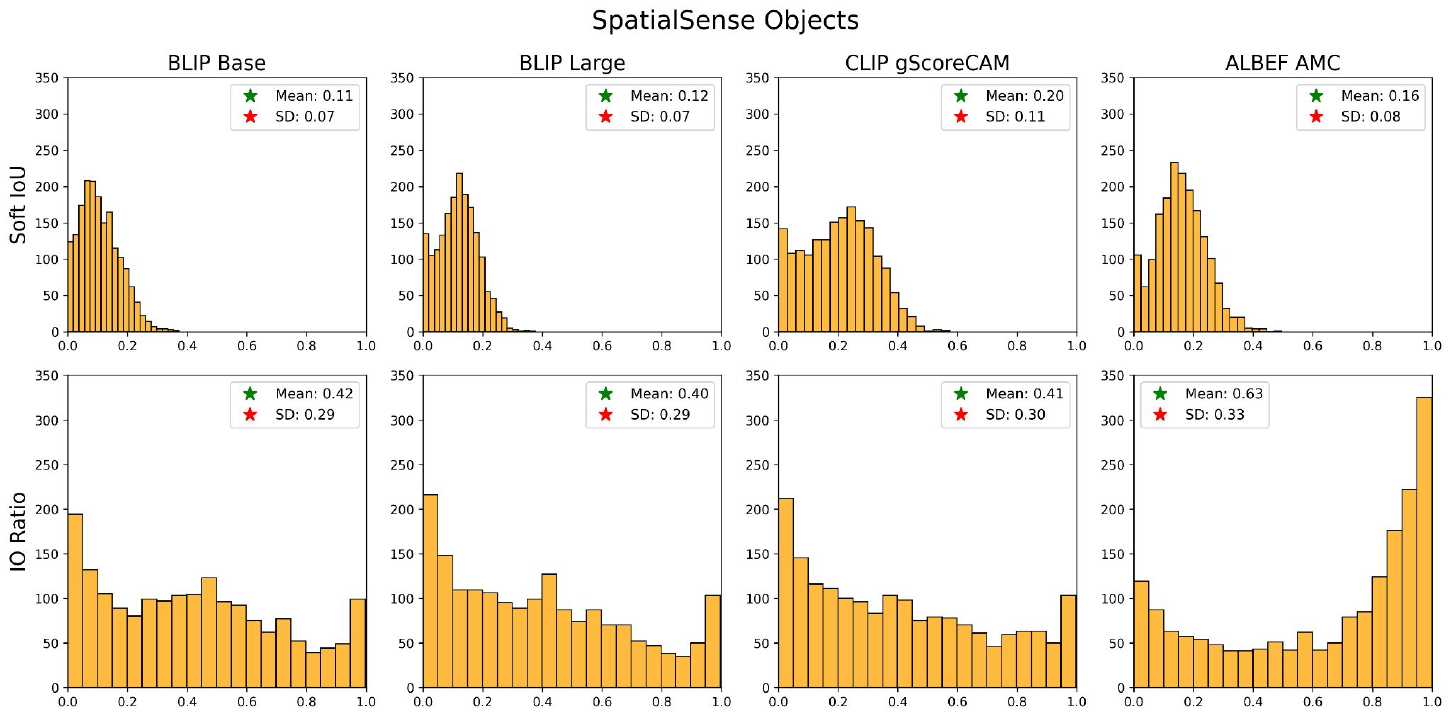}
   \caption{SpatialSense \textbf{objects}.}
   \label{fig:spatialsense_objects_hist_viz}
\end{figure*}

Flickr30K Entities \textbf{test} is provided in Figure~\ref{fig:hist_viz_fk30_final}.\\
RefCOCO+ \textbf{testA} is provided in Figure~\ref{fig:refCOCOPlus_testA_hist_viz}.\\
RefCOCO+ \textbf{testB} is provided in Figure~\ref{fig:refCOCOPlus_testB_hist_viz}.\\
SpatialSense \textbf{triplets} is provided in Figure~\ref{fig:spatialsense_triplets_hist_viz}.\\
SpatialSense \textbf{subjects} is provided in Figure~\ref{fig:spatialsense_subjects_hist_viz}.\\
SpatialSense \textbf{objects} is provided in Figure~\ref{fig:spatialsense_objects_hist_viz}.\\

\end{document}